\documentclass[conference]{IEEEtran}
\IEEEoverridecommandlockouts

\usepackage{natbib}
\usepackage{amsmath,amssymb,amsfonts}
\usepackage{algorithmic}
\usepackage{graphicx}
\usepackage{textcomp}
\usepackage{natbib}
\usepackage{csquotes}
\usepackage{hyperref}
\usepackage{threeparttable}
\usepackage{booktabs}
\usepackage{xcolor}
\usepackage{gensymb}
\usepackage{colortbl}
\usepackage{placeins}
\def\BibTeX{{\rm B\kern-.05em{\sc i\kern-.025em b}\kern-.08em
    T\kern-.1667em\lower.7ex\hbox{E}\kern-.125emX}}
\begin{document}


\title{Material-informed Gaussian Splatting for 3D World Reconstruction in a Digital Twin}

\author{
Andy Huynh$^{1,2*}$, João Malheiro Silva$^{1*}$, Holger Caesar$^{2}$, and Tong Duy Son$^{1}$%
\thanks{$^{*}$Equal contribution.}%
\thanks{$^{1}$Siemens Digital Industries Software, 3001 Leuven, Belgium. Email: \{silva.joao, son.tong\}@siemens.com}%
\thanks{$^{2}$Dept. of Cognitive Robotics, Delft University of Technology, 2628 CD Delft, The Netherlands. Email: \{a.huynh, H.Caesar\}@tudelft.nl}%
\thanks{Funded by the European Union. Views and opinions expressed are however those of the author(s) only and do not necessarily reflect those of the European Union or the European Health and Digital Executive Agency (HADEA). Neither the European Union nor the granting authority can be held responsible for them. This work is part of the ROBUSTIFAI (grant agreement No. 101212818) and SYNERGIES (grant agreement No. 101146542) projects, both funded by Horizon Europe. The work also benefited from VLAIO-funded projects SATISFY.AI and STREAMS.}%
}

\maketitle


\begin{abstract}
3D reconstruction for Digital Twins often relies on LiDAR-based methods, which provide 
accurate geometry but lack the semantics and textures naturally captured by cameras. 
Traditional LiDAR-camera fusion approaches require complex calibration and still struggle 
with certain materials like glass, which are visible in images but poorly represented in 
point clouds. We propose a camera-only pipeline that reconstructs scenes using 3D Gaussian 
Splatting from multi-view images, extracts semantic material masks via vision models, 
converts Gaussian representations to mesh surfaces with projected material labels, and 
assigns physics-based material properties for accurate sensor simulation in modern graphics 
engines and simulators. This approach combines 
photorealistic reconstruction with physics-based material assignment, providing sensor 
simulation fidelity comparable to LiDAR-camera fusion while eliminating hardware complexity 
and calibration requirements. We validate our camera-only method using an internal dataset 
from an instrumented test vehicle, leveraging LiDAR as ground truth for reflectivity 
validation alongside image similarity metrics.
\end{abstract}


\begin{figure*}[!t]
    \centering
    \includegraphics[width=1\linewidth,height=\paperheight,keepaspectratio]{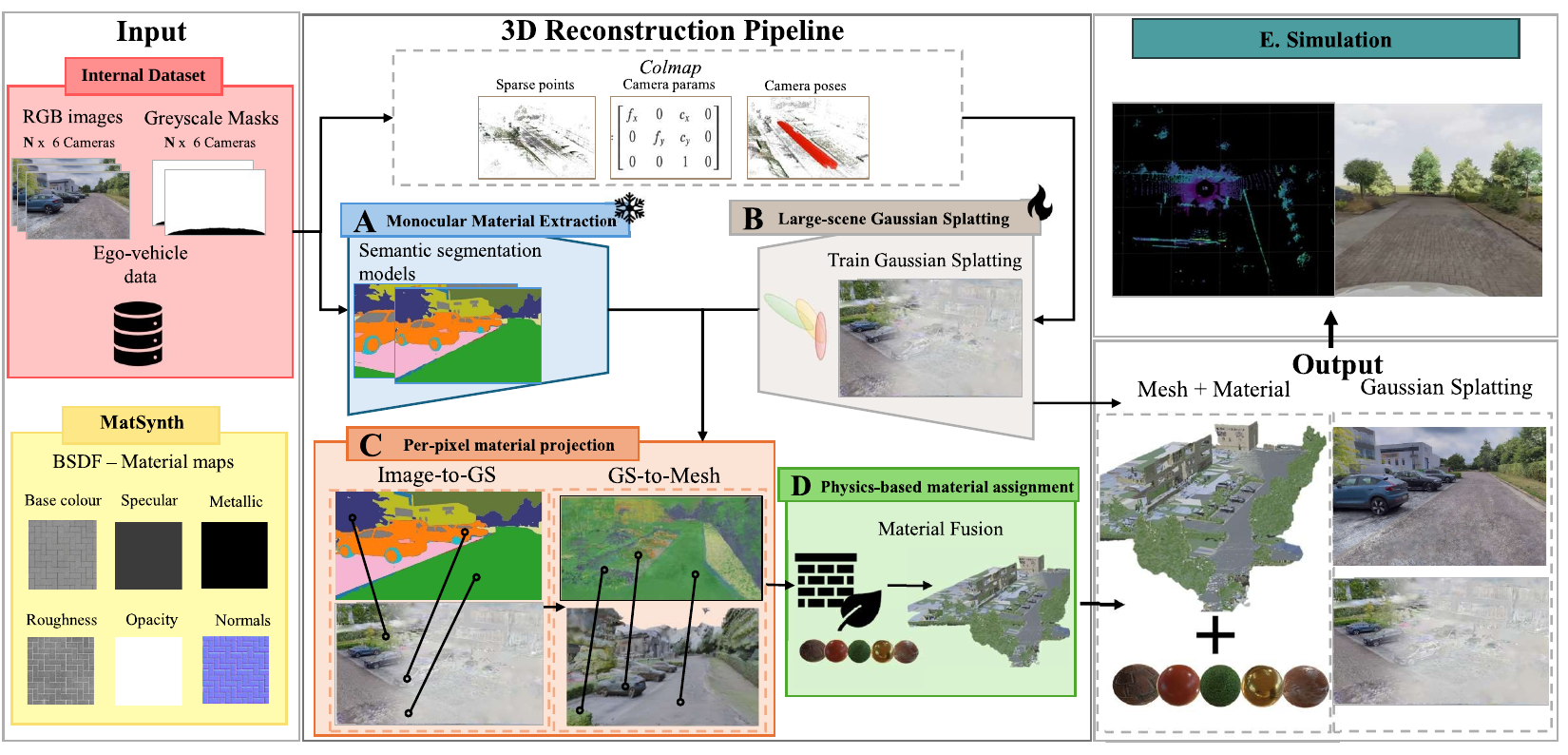}
    \caption{Overview of our camera-only reconstruction pipeline. From RGB images, 
    we: (A) extract semantic material masks, (B) reconstruct the 3D scene, 
    (C) project material labels onto mesh surfaces, (D) assign physics-based 
    materials, and (E) validate through sensor simulation.}
    \label{fig:3d_rec_pipeline}
\end{figure*}
\section{Introduction}
Digital Twins advance sensor technologies by enabling safe validation of high-risk 
scenarios and precise environmental control prior to real-world deployment. Modern 
Advanced Driver Assistance Systems (ADAS) and Artificial Intelligence (AI) systems 
rely on sophisticated algorithms and complex multimodal sensor setups, often including 
LiDAR sensors and cameras. Achieving accurate translation of real-world scenarios 
to simulation—where virtual sensors exhibit identical behavior to their physical 
counterparts—requires 3D world reconstruction that captures spatial and geometric 
features to ensure realistic simulation and synthetic sensor data consistent with 
reality.

Previous LiDAR-based approaches~\cite{Zaalberg2024, Manivasagam} achieve 
accurate geometry but lack camera-captured texture information. Moreover, LiDAR struggles with transparent and reflective materials like glass and metal, 
readily captured in camera images. While incorporating material properties into LiDAR-based reconstruction 
has been shown to improve sensor simulation accuracy~\cite{muckenhuber2020automotive}, 
such approaches still require LiDAR hardware and do not address texture 
acquisition. Combining LiDAR with cameras to capture both geometry and appearance 
requires complex sensor synchronization and calibration procedures, yet still 
produces limited texture quality under sparse-view conditions, as traditional 
reconstruction methods struggle to generate photorealistic results.

Recent advancements in novel view synthesis have introduced 3D Gaussian 
Splatting~\cite{kerbl20233dgaussiansplattingrealtime}, an efficient technique 
providing fast, differentiable rendering and high-quality reconstructions from 
sparse input views. Recent works like MILo~\cite{guédon2025milomeshintheloopgaussiansplatting} extract 
geometric surfaces from Gaussian Splatting using only multi-view images, enabling 
hybrid representations combining photorealistic visualization with physics-based 
simulation in graphics engines. These camera-only methods provide an alternative to LiDAR hardware, avoiding 
intricate sensor synchronization and demanding calibration procedures while 
delivering superior texture quality.

Building on these observations, we propose a camera-only pipeline that combines 
photorealistic Gaussian Splatting with physics-based material assignment 
(Fig.~\ref{fig:3d_rec_pipeline}). This approach enables accurate sensor simulation 
by integrating semantic material information into visually realistic reconstructions.

Our main \textbf{contributions} are:
\begin{itemize}
    \item \textbf{(1)} An automated method for projecting 2D semantic material 
    masks onto 3D mesh surfaces through Gaussian Splatting-based reconstruction, 
    enabling accurate physics-based LiDAR reflectivity simulation.
    
    \item \textbf{(2)} A modular, camera-only pipeline integrating Gaussian 
    Splatting for photorealistic geometry reconstruction with automated mesh-based 
    material assignment.

    \item \textbf{(3)} Comprehensive evaluation demonstrating sensor simulation 
    accuracy comparable to LiDAR-camera fusion, validated through LiDAR reflectivity 
    and rendering quality analysis.
\end{itemize}

While our pipeline requires only camera input, we validate it using LiDAR ground 
truth from an instrumented test vehicle.


\section{Related Work}\label{sec:related_works}

\subsection{Novel View Synthesis}
Traditional 3D reconstruction methods use explicit representations. LiDAR-based 
methods~\cite{huang2023neural} excel at geometric accuracy but struggle 
with fine visual details. Camera-based alternatives~\cite{Schonberger2016} 
offer richer visual information but produce noisier geometry and face challenges with 
transparent and reflective surfaces~\cite{remondino2023critical}. Neural Radiance Fields 
(NeRF)~\cite{mildenhall2020nerfrepresentingscenesneural} enabled photorealistic novel 
view synthesis from camera-only input but suffers from high computational overhead and 
lacks discrete surfaces needed for graphics engines.

3D Gaussian Splatting~\cite{kerbl20233dgaussiansplattingrealtime} addressed these 
limitations through an explicit point-based representation enabling real-time rendering 
with photorealistic quality. Subsequent refinements have improved geometric accuracy, 
transparency handling, and shading~\cite{Huang_2024,zhang2024quadraticgaussiansplattingefficient,wu20253dgut}. 
However, these methods remain primarily oriented toward small, object-centric captures 
and struggle with large-scale outdoor scenes.

\subsection{Large-Scale Scene Reconstruction}
To scale Gaussian Splatting to large-scale outdoor environments, recent methods have 
adopted scene partitioning strategies~\cite{tancik2022blocknerfscalablelargescene}, 
dividing scenes into spatially consistent blocks for parallel reconstruction~\cite{liu2025citygaussianv2efficientgeometricallyaccurate,h3dgs}. 
Many employ Level-of-Detail (LoD) techniques, allocating higher resolution to nearby 
regions to improve rendering speed. Alternative strategies include temporal 
segmentation~\cite{cui2024streetsurfgsscalableurbanstreet} and distance-based 
decomposition~\cite{shi2024dhgsdecoupledhybridgaussian}. However, these methods often 
require manual parameter tuning and lack robust surface representations needed for 
simulation engines.

\subsection{Surface Extraction from Gaussian Primitives}
While Gaussian Splatting enables high-quality rendering, simulation engines require 
explicit mesh representations. Methods such as~\cite{guedon2023sugar} use Poisson 
reconstruction, while Signed Distance Field (SDF)-based 
approaches~\cite{chen2025neusgneuralimplicitsurface} require dense multi-view data. 
Recent work~\cite{guédon2025milomeshintheloopgaussiansplatting} integrates mesh 
extraction directly into training via Delaunay triangulation, enabling 
joint optimization of appearance and geometry. However, these methods are primarily 
designed for controlled, object-centric scenarios and lack robustness for large-scale 
outdoor environments.

\subsection{Material Extraction}\label{sec2:material_extraction}
Beyond extracting accurate geometry, material properties are essential for realistic 
simulation. Image-based methods classify materials through 
semantic segmentation~\cite{cai2022rgb}. 
Multi-modal approaches~\cite{Liang_2022_CVPR} and large-scale 
datasets~\cite{upchurch2022densematerialsegmentationdataset} have improved accuracy. 
While LiDAR-based methods~\cite{muckenhuber2020automotive, viswanath2024reflectivityneedadvancinglidar} measure physical properties directly, camera-only approaches capture 
materials that LiDAR struggles with, such as transparent and reflective surfaces. 
However, transferring 2D material segmentation to 3D reconstructed surfaces remains 
challenging for camera-only Gaussian Splatting pipelines, where material labels must 
be accurately projected from multi-view observations onto extracted mesh geometry. 
Our work addresses this gap by introducing an automated projection method that enables 
physics-based sensor simulation.
\begin{figure*}[t]
    \centering
    \includegraphics[width=0.9\linewidth,height=\paperheight,keepaspectratio]{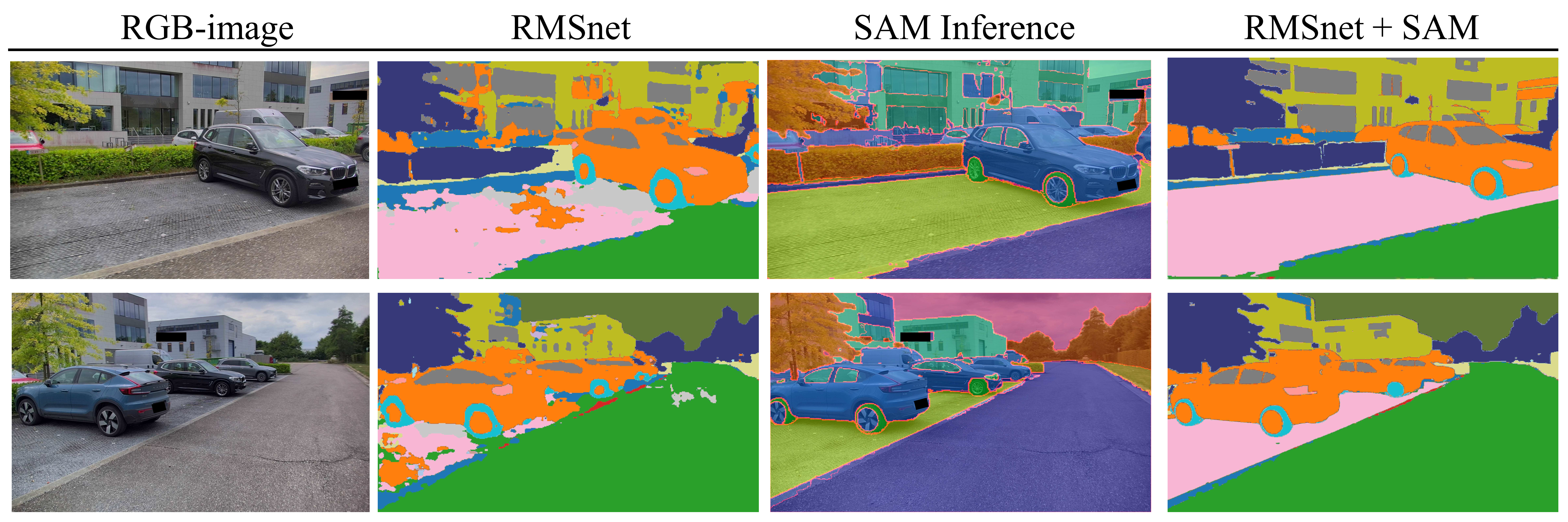}
    \caption{Shape-aware material refinement. From left to right: input RGB image, 
    RMSNet predictions, FastSAM object boundaries, and final result after majority 
    voting. The combination produces consistent material labels with sharp edges. 
    Example from our internal dataset.}
    \label{fig:sam_rmsnet}
\end{figure*}
\subsection{Physics-Based Materials for Rendering}
Once material labels are assigned to 3D surfaces, they must be represented in a format 
compatible with physics-based rendering engines. Modern graphics engines simulate surface 
interactions through the Bidirectional Reflectance Distribution Function (BRDF)~\cite{murat2018survey}. 
Recent work integrates material properties directly into 3D Gaussians~\cite{
ye2025largematerialgaussianmodel, yao2025reflectivegaussiansplatting} focusing on rendering, 
while mesh-compatible approaches~\cite{Xiong_2025_CVPR} enable seamless integration with 
standard simulation engines. However, bridging the gap between semantic material labels from 
2D segmentation and physically accurate material parameters for physics-based sensor 
simulation in Digital Twin environments remains an open challenge.


\section{Method}
Our pipeline reconstructs large-scale outdoor scenes from multi-view RGB images 
and assigns physics-based materials for accurate sensor simulation. The approach 
is \textit{decoupled}, separating geometric reconstruction from material 
assignment, and employs a \textit{hybrid representation}, combining photorealistic 
Gaussian Splatting with explicit mesh geometry. As illustrated in Fig.~\ref{fig:3d_rec_pipeline}, 
the pipeline consists of five stages: monocular material extraction 
(Sec.~\ref{sec:material_extraction}), Gaussian Splatting reconstruction 
(Sec.~\ref{sec:gs}), per-pixel material projection from 2D to 3D 
(Sec.~\ref{sec:projection}), physics-based material assignment (Sec.~\ref{sec:pbr}), 
and simulation validation (Sec.~\ref{sec:simulation}).

\subsection{Monocular Material Extraction}\label{sec:material_extraction}

We extract per-pixel material labels from RGB images using a two-stage approach: 
texture-based material segmentation followed by shape-aware refinement. For the 
first stage, we adopt RMSNet~\cite{cai2022rgb} due to its strong performance on 
autonomous driving datasets and its ability to operate on RGB images alone, 
without requiring LiDAR or infrared sensors. It distinguishes classes such as 
asphalt, concrete, glass, and metal.

Material regions often exhibit fragmented distributions lacking prominent shape cues. 
We employ FastSAM~\cite{zhao2023fastsegment} to identify instances, producing 
pixel-wise boundaries that we post-process to remove overlaps before overlaying 
with RMSNet predictions.

For each region, we apply majority voting: counting material class 
labels within its boundary and assigning the most frequent label to all enclosed 
pixels. This strategy ensures consistent material assignment while preserving 
sharp boundaries, as illustrated in Fig.~\ref{fig:sam_rmsnet}. The result is a 
set of \textit{n} shape-aware material masks, where each pixel inherits its 
class according to the FastSAM-derived segmentation.

\subsection{Large-Scale Gaussian Splatting}\label{sec:gs}

Our 3D reconstruction combines photorealistic rendering with explicit geometry 
through a hybrid approach. We first apply COLMAP~\cite{Schonberger2016} to generate 
a sparse point cloud from multi-view RGB images, providing GPS-derived camera poses, 
pre-calibrated intrinsics, and extrinsics as priors. We mask the moving ego-vehicle in all input images to avoid artifacts.

From the COLMAP output, we train two complementary models. For photorealistic visualization, we use 
H3DGS~\cite{h3dgs}, selected for its superior visual quality on autonomous driving 
datasets over alternatives like CityGaussianV2~\cite{liu2025citygaussianv2efficientgeometricallyaccurate}. H3DGS 
constructs a global scaffold and subdivides the scene into chunks. However, H3DGS's 
hierarchical LoD representation uses a proprietary \texttt{.hier} format, and its 
anti-aliasing mechanism~\cite{yu2023mip} is not supported by standard 3DGS rendering 
tools that expect the conventional \texttt{.ply} format. To maximize compatibility 
with existing 3DGS viewers and downstream processing tools, we therefore disable 
these features and perform a custom merging of the chunks.

For mesh geometry and semantic labeling, we employ MiLO~\cite{guédon2025milomeshintheloopgaussiansplatting}, 
which provides geometrically accurate mesh extraction required for physics-based 
simulation. Trained on the same COLMAP sparse point cloud, MiLO produces both a 
3D Gaussian model and an explicit mesh. The MiLO Gaussian model initializes our 
semantic Gaussian Splatting for material projection (Sec.~\ref{sec:projection}), 
while the mesh provides the geometric basis for physics-based simulation 
(Sec.~\ref{sec:simulation}). Both H3DGS and MiLO are trained with monocular depth 
supervision~\cite{depth_anything_v2} following their respective default configurations, and 
both reconstructions share the same coordinate system, ensuring alignment between 
appearance and geometry.

\begin{figure*}[!htp]
    \centering
    \includegraphics[width=0.85\linewidth]{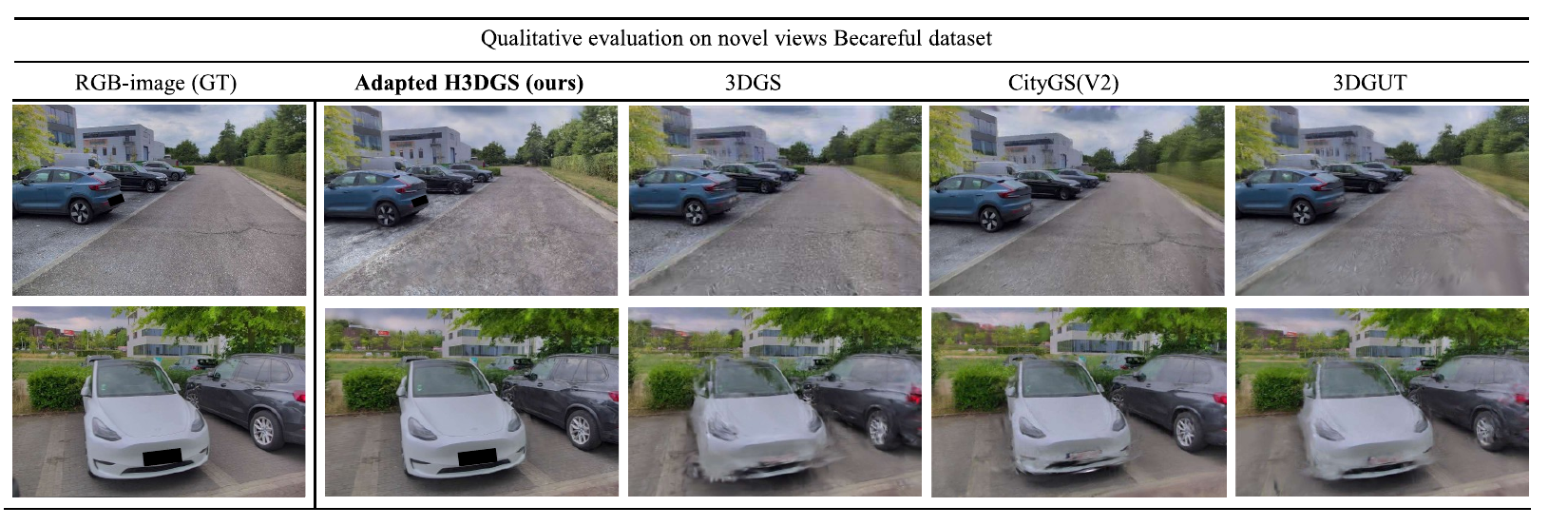}
    \caption{Qualitative comparison of novel view rendering. From left to right: ground 
    truth RGB, our adapted H3DGS model, 3DGS, CityGaussianV2, and 3DGUT. Our model 
    achieves competitive visual quality comparable to state-of-the-art baselines.}
    \label{fig:qualitative_comparison}
\end{figure*}

\subsection{Per-Pixel Material Projection}\label{sec:projection}

This module projects the 2D material masks from Sec.~\ref{sec:material_extraction} 
onto the MiLO Gaussian representation, then transfers the labels to the mesh. 
Unlike previous pipelines relying on camera-to-LiDAR transformations, our Gaussian 
model is natively aligned in world coordinates. We employ differentiable 
rasterization to render Gaussians to 2D, maintaining 
correspondence between 3D Gaussian positions and 2D pixel coordinates. When 
multiple Gaussians project to the same pixel, depth sorting during alpha blending 
assigns the label of the closest Gaussian.

Inconsistencies across views, such as static objects receiving different labels, 
are resolved using SegAnyGS~\cite{cen2025segment3dgaussians}, which enforces 
consistent assignment of our material masks to the Gaussians. We apply majority 
voting across overlaid masks to ensure label consistency. Each material-labeled 
Gaussian is assigned to its nearest mesh triangle using K-Nearest Neighbors 
(KNN), and the triangle inherits the Gaussian's 
label, producing a per-triangle material mesh as the final output.

\subsection{Physics-Based Material Assignment}\label{sec:pbr}

We assign physics-based material properties to the mesh using the Principled 
BSDF shader following the Disney BSDF standard~\cite{burley2012physically}, 
ensuring compatibility with graphics engines such as Blender, Unreal Engine, 
and simulation platforms like Simcenter Prescan. The Principled BSDF is parameterized by physically meaningful properties including base color, specular, metallic factor, roughness, opacity, and surface normals.

From the 20 material classes detected by RMSNet (Sec.~\ref{sec:material_extraction}), 
we selected a subset commonly encountered in urban driving environments: 
glass, brick and ceramics, concrete, asphalt, vegetation (e.g., grass and 
leaves), metals, plastics, gravel, tree trunks, and rubber. For each 
category, PBR textures were sourced from the Matsynth dataset~\cite{Vecchio_2024}, 
which provides over 4,000 physically-based materials across 14 categories, 
and matched to Prescan's laboratory-tested material database to ensure 
validated reflectivity properties for LiDAR simulation. Within each category, 
textures were balanced across key parameters (base color, roughness, metallic 
factor, clear coat). For each material class from Sec.~\ref{sec:material_extraction}, 
we apply the selected PBR textures to the corresponding mesh triangles, 
producing a fully textured mesh directly usable by standard graphics engines 
and physics-based sensor simulation frameworks.

\subsection{Simulation Validation}\label{sec:simulation}

To validate our material-informed reconstruction pipeline, we integrate the 
reconstructed scenes into Simcenter Prescan to recreate realistic traffic 
scenarios with physics-based sensor simulation. The material-labeled mesh from 
Sec.~\ref{sec:pbr} is imported into Prescan, where laboratory-measured PBR 
material properties are assigned to each triangle based on its material label. 
This enables accurate interaction with physics-based sensors, particularly LiDAR, 
where surface reflectivity depends on material composition.

We replicate the original sensor configuration from our instrumented test vehicle, 
including LiDAR specifications and mounting positions. The ego-vehicle trajectory 
is extracted and imported via the Simcenter Prescan MATLAB API, ensuring the virtual 
sensor follows the exact path and poses as during real-world data capture. The 
simulation generates synthetic LiDAR point clouds with material-specific reflectivity 
values for each point.

We validate our camera-only reconstruction by comparing these simulated LiDAR 
responses against real-world LiDAR measurements from the instrumented test vehicle. 
This demonstrates that our pipeline produces 3D models suitable for physics-based 
sensor simulation, providing a practical alternative to traditional LiDAR-based 
reconstruction workflows while achieving comparable sensor simulation accuracy.


\FloatBarrier
\begin{table*}[!t]
    \centering
    \begin{threeparttable}
        \caption{Image similarity metrics (PSNR $\uparrow$, SSIM $\uparrow$, LPIPS $\downarrow$) evaluated on novel views for all baselines.}
        \label{tab:image-similarity-metrics}
        \renewcommand{\arraystretch}{1.3}
        \begin{tabular}{l ccc| ccc| ccc| ccc}
            \toprule
            & \multicolumn{3}{c|}{\textbf{Adapted H3DGS (Ours)}}
            & \multicolumn{3}{c|}{3DGS}
            & \multicolumn{3}{c|}{CityGaussianV2}
            & \multicolumn{3}{c}{3DGUT} \\
            & \textbf{PSNR$\uparrow$} & \textbf{SSIM$\uparrow$} & \textbf{LPIPS$\downarrow$}
            & \textbf{PSNR$\uparrow$} & \textbf{SSIM$\uparrow$} & \textbf{LPIPS$\downarrow$}
            & \textbf{PSNR$\uparrow$} & \textbf{SSIM$\uparrow$} & \textbf{LPIPS$\downarrow$}
            & \textbf{PSNR$\uparrow$} & \textbf{SSIM$\uparrow$} & \textbf{LPIPS$\downarrow$} \\
            \midrule
            \textbf{Scene 1} & 16.88 & 0.299 & \cellcolor{yellow!30}0.451 & 19.13 & \cellcolor{yellow!30}0.476 & \cellcolor{red!30}0.438 & \cellcolor{red!30}20.36 & \cellcolor{red!30}0.486 & 0.480 & \cellcolor{yellow!30}19.97 & 0.465 & 0.517 \\
            \textbf{Scene 2} & 17.86 & 0.358 & \cellcolor{red!30}0.432 & 19.81 & \cellcolor{yellow!30}0.525 & \cellcolor{yellow!30}0.474 & \cellcolor{yellow!30}20.79 & \cellcolor{red!30}0.526 & 0.522 & \cellcolor{red!30}21.16 & 0.505 & 0.512 \\
            \textbf{Scene 3} & 18.35 & 0.375 & \cellcolor{red!30}0.437 & 19.48 & \cellcolor{yellow!30}0.515 & \cellcolor{yellow!30}0.492 & \cellcolor{yellow!30}20.42 & \cellcolor{red!30}0.524 & 0.540 & \cellcolor{red!30}21.10 & 0.502 & 0.557 \\
            \textbf{Scene 4} & 19.03 & 0.504 & \cellcolor{red!30}0.439 & 20.88 & 0.659 & \cellcolor{yellow!30}0.450 & \cellcolor{yellow!30}21.15 & \cellcolor{red!30}0.676 & 0.483 & \cellcolor{red!30}21.85 & \cellcolor{yellow!30}0.666 & 0.526 \\
            \textbf{Scene 5} & 19.21 & 0.503 & \cellcolor{red!30}0.427 & 21.04 & 0.650 & \cellcolor{yellow!30}0.446 & \cellcolor{yellow!30}21.63 & \cellcolor{red!30}0.690 & 0.499 & \cellcolor{red!30}22.04 & \cellcolor{yellow!30}0.662 & 0.520 \\
            \midrule
            \textbf{Average} & \textbf{18.27} & \textbf{0.41} & \cellcolor{red!30}\textbf{0.44} & \textbf{20.07} & \textbf{0.57} & \cellcolor{yellow!30}\textbf{0.46} & \cellcolor{yellow!30}\textbf{20.87} & \cellcolor{red!30}\textbf{0.58} & \textbf{0.50} & \cellcolor{red!30}\textbf{21.22} & \cellcolor{yellow!30}\textbf{0.56} & \textbf{0.53} \\
            \bottomrule
        \end{tabular}
        \begin{tablenotes}
            \footnotesize
            \item Best performance highlighted in red, second-best in yellow.
        \end{tablenotes}
    \end{threeparttable}
\end{table*}

\section{Experiments}

We evaluate our camera-only reconstruction pipeline using our internal dataset. 
To assess whether camera-only reconstruction achieves sensor simulation accuracy 
comparable to LiDAR-based methods, we establish a LiDAR-camera fusion baseline 
that combines LiDAR-based mesh reconstruction (NKSR~\cite{huang2023neural}) with 
camera-derived material labels. This baseline provides geometric accuracy through 
direct depth measurements while incorporating semantic material information from 
camera images.

We conduct two complementary evaluations with different baselines: (1) \textit{visual 
quality}, where we compare our Gaussian Splatting reconstruction against state-of-the-art 
camera-only methods (3DGS~\cite{kerbl20233dgaussiansplattingrealtime}, 
3DGUT~\cite{wu20253dgut}, CityGaussianV2~\cite{liu2025citygaussianv2efficientgeometricallyaccurate}) 
using image similarity metrics (PSNR, SSIM, LPIPS), and (2) \textit{sensor simulation 
accuracy}, where we compare our camera-only approach against the LiDAR-camera fusion 
baseline by evaluating synthetic LiDAR reflectivity against real-world ground truth.

\subsection{Dataset}
We collected data using an instrumented test vehicle equipped with an Ouster OS1-128 
LiDAR~\cite{ousterOS1Specs} (128 channels, 20 Hz, 100 Hz IMU), six surround-view cameras with Sony 
IMX490 sensors (2880×1860 resolution, one rear with fisheye lens for 360° coverage), 
and a Septentrio AsteRx GNSS/RTK system (20 Hz). The LiDAR provides hardware triggers 
to synchronize all cameras at 20 Hz. Five static road scenes were captured near our 
company facilities in Leuven, Belgium, each approximately eight seconds at 20-30 km/h. Data collection on company-controlled roads avoided 
GDPR compliance issues and ensured controlled validation conditions.

We focus on static scene reconstruction as this aligns with Digital Twin simulation 
workflows. Modern simulation platforms (e.g., Simcenter Prescan, CARLA) provide 
realistic dynamic actor models with controllable behaviors, making the challenge the 
creation of photorealistic static environments (roads, buildings, vegetation, 
infrastructure). Our method automates this process, providing realistic backgrounds 
for simulation scenarios. Dynamic actors in our recordings are masked out during 
reconstruction. While our detection pipeline enables trajectory replication for 
re-insertion as simulation assets, this falls outside the scope of this paper.

Scenes primarily consist of company parking lots and nearby access roads, capturing 
typical automotive testing environments with diverse material challenges including 
asphalt, concrete, glass facades, metallic surfaces, extensive vegetation (trees, 
hedges, grass), road markings, and signage. Recordings under varying weather 
conditions (clear to partly cloudy) capture natural lighting variations. We reserve 
50 novel viewpoints per scene for evaluation.

\subsection{Baseline}

\noindent\textbf{Visual quality:} We compare our Gaussian Splatting reconstruction 
against three state-of-the-art camera-only methods: 3DGS~\cite{kerbl20233dgaussiansplattingrealtime}, 
3DGUT~\cite{wu20253dgut}, and CityGaussianV2~\cite{liu2025citygaussianv2efficientgeometricallyaccurate}. 
All methods are trained on the same five scenes and evaluated on 50 uniformly sampled 
novel viewpoints per scene using PSNR, SSIM, and LPIPS.

\noindent\textbf{Sensor simulation accuracy:} We compare our camera-only reconstruction 
against a LiDAR-camera fusion baseline by simulating both environments in Simcenter 
Prescan with identical PBR material properties and ego-vehicle trajectories. The 
LiDAR-based baseline uses NKSR~\cite{huang2023neural} for mesh reconstruction from 
LiDAR point clouds, combined with the same camera-derived material labels as our 
camera-only approach. This controlled setup minimizes external variables, allowing 
direct comparison of geometric reconstruction quality and its impact on reflectivity 
accuracy.
\begin{figure*}[!htbp]
    \centering
    \includegraphics[width=0.85\linewidth,height=\paperheight,keepaspectratio]{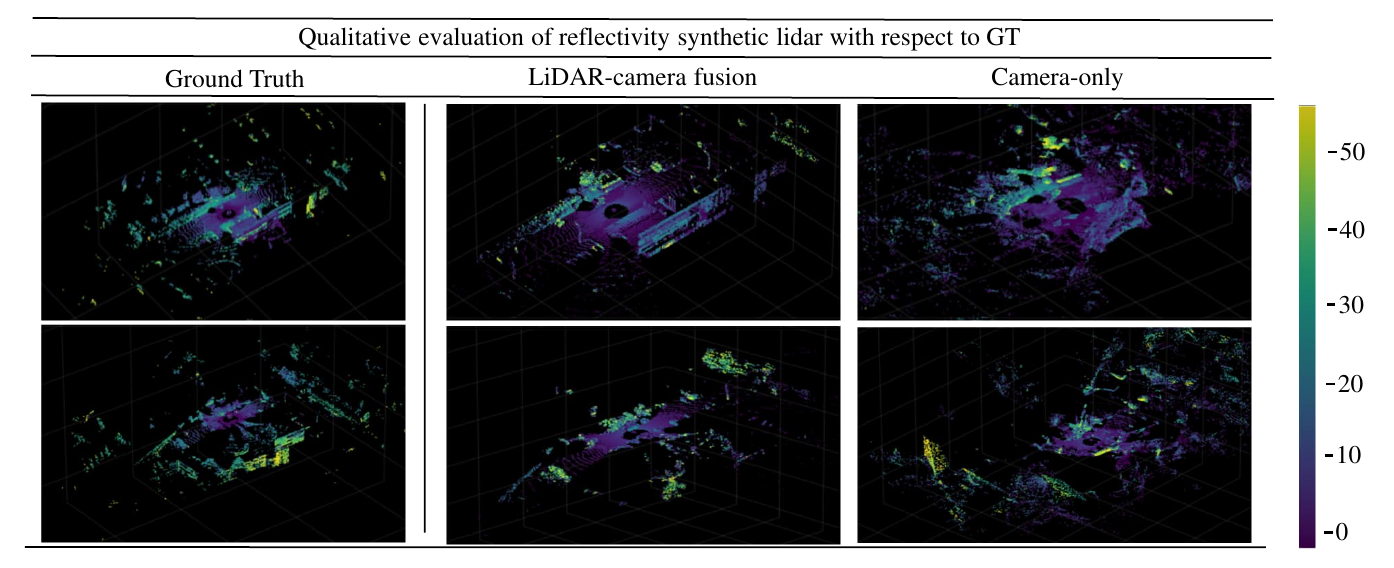}
    \caption{From left to right: 
    real-world LiDAR ground truth from our instrumented vehicle, LiDAR-camera fusion 
    baseline simulated in Prescan, and our camera-only reconstruction simulated in Prescan.}
    \label{fig:lidar_reflectivity_comparison}
\end{figure*}
\subsection{Implementation Details}

\noindent\textbf{Scene reconstruction:} We train all Gaussian Splatting methods 
using their default hyperparameters as specified in their respective repositories. 
For both H3DGS and MiLO, we enable monocular depth supervision using Depth Anything 
V2~\cite{depth_anything_v2}. Material assignment follows the pipeline described in 
Sec.~\ref{sec:material_extraction}--\ref{sec:pbr}, projecting semantic labels onto 
the reconstructed geometry and mapping them to PBR material properties.

\noindent\textbf{LiDAR simulation:} We configure the Simcenter Prescan LiDAR sensor 
according to the specifications of the Ouster OS1-128~\cite{ousterOS1Specs} used 
in our dataset. The physics-based sensor returns both Cartesian coordinates of 
detected points and a power value representing reflected signal strength. We 
normalize this power output for range and incidence angle to obtain accurate 
surface reflectivity values, following established calibration 
procedures~\cite{hermidas_phillips_physlidarpowercal}.

\noindent\textbf{Hardware:} All experiments were conducted on a Dell Precision 7680 
equipped with 64GB RAM and an NVIDIA RTX A4000 ADA Laptop GPU. Sensor validation 
was performed using Simcenter Prescan 2411, with scenario control via the 
Prescan MATLAB API (R2023b).

\subsection{Metrics}\label{e4:metrics}

We employ complementary metrics for visual quality and sensor simulation accuracy:

\noindent\textbf{Visual quality:} PSNR~\cite{gonzalez2008digital} (pixel-level 
fidelity), SSIM~\cite{1284395} (structural similarity), and 
LPIPS~\cite{zhang2018unreasonableeffectivenessdeepfeatures} (perceptual quality). 
Higher is better for PSNR and SSIM; lower is better for LPIPS.

\noindent\textbf{LiDAR simulation accuracy:} Mean Absolute Error (MAE) and median 
error quantify reflectivity prediction error between synthetic and real-world 
measurements, with median providing robustness against outliers.

\subsection{Visual Quality Evaluation}\label{sec:visual_quality}

We evaluate our adapted H3DGS model against three state-of-the-art Gaussian 
Splatting methods by comparing novel view renderings across all five scenes.

\noindent\textbf{Qualitative analysis.} Figure~\ref{fig:qualitative_comparison} 
demonstrates that our adapted model achieves competitive visual quality with 
minimal deviation from ground truth. The model effectively fills masked ego-vehicle 
regions and produces cleaner reconstructions by filtering sensor noise in the 
ground truth. However, limitations emerge when reconstructing highly reflective 
and transparent surfaces, occasionally resulting in misalignments with ground 
truth. Rendering quality is highest for front-facing cameras, while side cameras 
exhibit reduced quality due to motion blur from the vehicle's movement. Scenes 
with dense object distributions and occlusions also reduce reconstruction quality.

\noindent\textbf{Quantitative analysis.} Table~\ref{tab:image-similarity-metrics} 
presents PSNR, SSIM, and LPIPS metrics across all methods and scenes. Our adapted 
H3DGS model is used solely for photorealistic visualization, while MiLO provides 
geometry for simulation (Sec.~\ref{sec:gs}). The model achieves lower PSNR 
(18.27 vs. 20.07--21.22) and SSIM (0.41 vs. 0.56--0.58) compared to baselines, 
but the best LPIPS score (0.44 vs. 0.46--0.53), outperforming all methods on 
four out of five scenes. Since LPIPS correlates strongly with human perceptual 
similarity~\cite{zhang2018unreasonableeffectivenessdeepfeatures}, this suggests 
our reconstruction maintains perceptual quality despite lower pixel-level accuracy.

The degradation stems from disabling H3DGS's hierarchical optimization and 
anti-aliasing mechanism~\cite{yu2023mip}. As noted by the H3DGS authors, 
hierarchies operating at different scales require correct anti-aliasing~\cite{h3dgs}. 
Anti-aliasing alone improves 3DGS quality by +4.56 dB PSNR and +0.070 SSIM~\cite{yu2023mip}; 
our observed degradation (-1.80 dB PSNR, -0.15 SSIM) is consistent with these 
modifications. The superior LPIPS score confirms these changes preserve perceptual 
quality while ensuring compatibility with standard rendering tools.

\subsection{LiDAR Simulation Accuracy}\label{sec:lidar_simulation}

We evaluate our camera-only Gaussian Splatting reconstruction against the 
LiDAR-camera fusion baseline by measuring reflectivity error relative to 
real-world ground truth.

\begin{table}[!ht]
    \centering
    \begin{threeparttable}
        \caption{LiDAR reflectivity prediction error (lower is better). Reflectivity 
        values normalized to 0--255 range.}
        \label{tab:reflectivity-metrics1}
        \renewcommand{\arraystretch}{1.2}
        \begin{tabular}{l cc cc}
            \toprule
            & \multicolumn{2}{c}{\textbf{Camera-only (Ours)}} & \multicolumn{2}{c}{\textbf{LiDAR-based}} \\
            \cmidrule(lr){2-3} \cmidrule(lr){4-5}
            & MAE$\downarrow$ & Median$\downarrow$ & MAE$\downarrow$ & Median$\downarrow$ \\
            \midrule
            Scene 1 & 10.91 & 7.16 & 11.73 & 7.84 \\
            Scene 2 & 9.37 & 5.54 & 10.17 & 6.70 \\
            Scene 3 & 11.03 & 6.91 & 10.47 & 6.87 \\
            Scene 4 & 9.32 & 5.77 & 9.41 & 6.02 \\
            Scene 5 & 8.98 & 6.71 & 8.76 & 6.36 \\
            \midrule
            \textbf{Average} & \textbf{10.05} & \textbf{6.48} & \textbf{10.14} & \textbf{6.78} \\
            \bottomrule
        \end{tabular}
        \begin{tablenotes}
            \footnotesize
            \item MAE: Mean Absolute Error. Our camera-only method achieves 
            comparable accuracy to the LiDAR-based baseline.
        \end{tablenotes}
    \end{threeparttable}
\end{table}

\noindent\textbf{Qualitative analysis.} Figure~\ref{fig:lidar_reflectivity_comparison} 
presents a visual comparison of LiDAR reflectivity predictions between our camera-only 
method, the LiDAR-camera fusion baseline, and real-world ground truth. Both methods 
consistently overestimate vegetation reflectivity due to: (1) temporal inconsistency 
from wind-induced vegetation movement violating static scene assumptions, (2) high 
intra-class material variance (leaves, bark, grass share one label but have different 
reflectivities), (3) idealized PBR materials not capturing weathered outdoor conditions, 
and (4) geometric complexity affecting surface normal estimation. These limitations 
impact both reconstruction approaches equally, indicating bottlenecks in the static 
scene assumption and material parameterization rather than reconstruction modality. 
Despite higher geometric noise in the camera-only method, both approaches produce 
comparable reflectivity patterns, validating our pipeline for sensor simulation.

\begin{figure}[!htp]
    \centering
    \includegraphics[width=1\columnwidth]{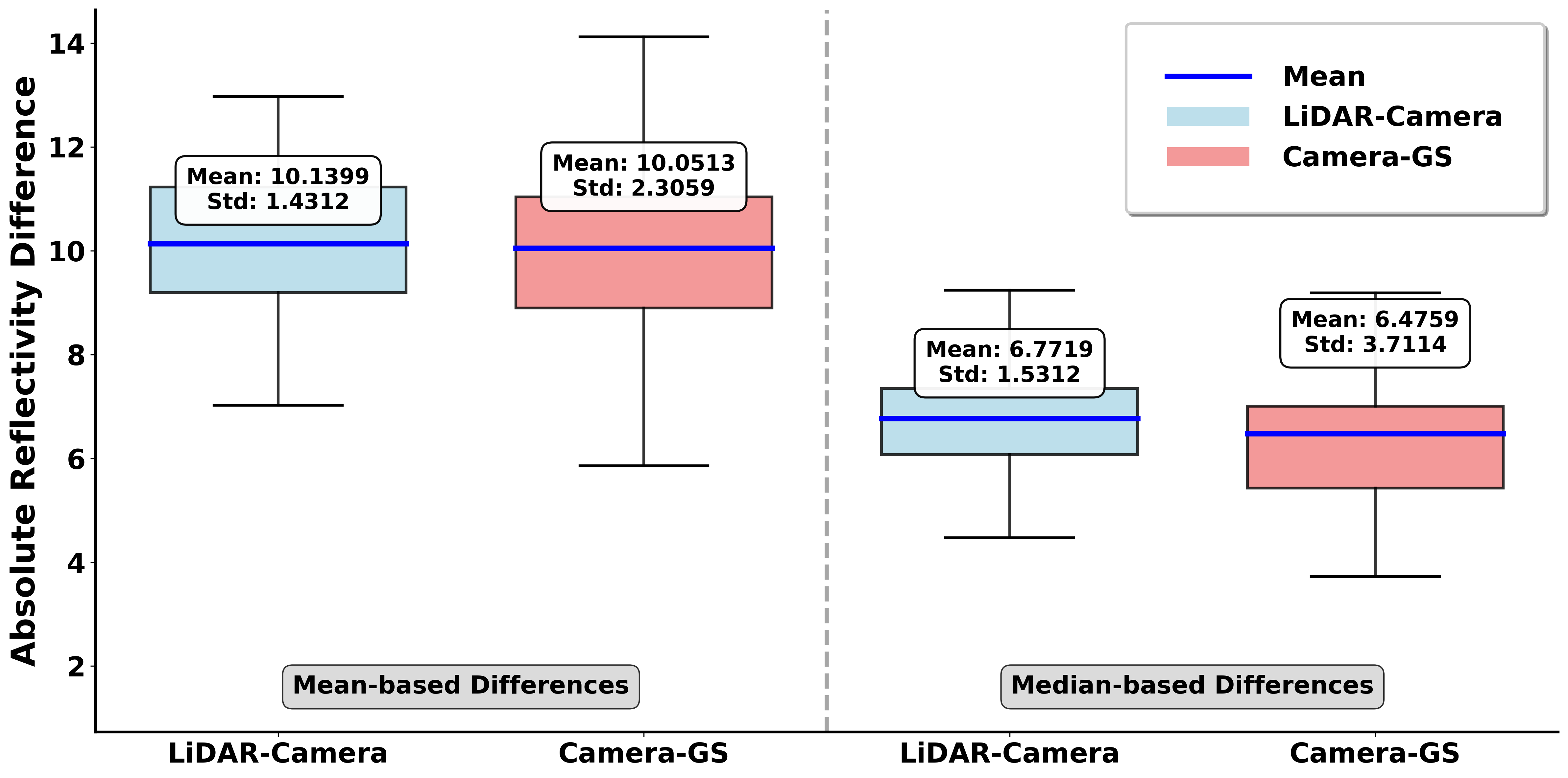}
    \caption{Distribution of absolute reflectivity errors for camera-only 
    (ours) and LiDAR-based reconstruction methods. Our camera-only approach 
    achieves comparable median error with slightly higher variability.}
    \label{fig:boxplot_representation}
\end{figure}

\noindent\textbf{Quantitative analysis.} Table~\ref{tab:reflectivity-metrics1} 
and Figure~\ref{fig:boxplot_representation} present reflectivity error metrics 
across all five scenes. Our camera-only method achieves comparable reflectivity 
accuracy to the LiDAR-camera fusion baseline, with mean absolute error 
(MAE: 10.05 vs. 10.14) and median error (6.48 vs. 6.78) showing minimal 
differences between the two approaches. This demonstrates that camera-only 
reconstruction can achieve sensor simulation accuracy comparable to LiDAR-based 
methods without requiring LiDAR hardware or complex sensor synchronization. 
The higher within-scene variability in camera-only results (visible in 
Figure~\ref{fig:boxplot_representation}) reflects the increased geometric 
noise inherent to MiLO's camera-based mesh extraction, which lacks the 
direct depth measurements provided by LiDAR. However, the comparable MAE 
and median errors demonstrate that this geometric uncertainty does not 
significantly degrade reflectivity simulation accuracy.

\subsection{Ablation Study}\label{sec:ablation}

We evaluate the impact of incorporating segmentation masks to improve material 
classification accuracy. As described in Sec.~\ref{sec:material_extraction}, 
we augment RMSNet's material predictions with shape cues from segmentation models.

\noindent\textbf{Experimental setup.} We quantitatively assess segmentation 
quality using the MCubes multi-modal dataset~\cite{Liang_2022_CVPR}, which contains 
500 material-annotated images. We compare three configurations: (1) RMSNet alone, 
(2) RMSNet + FastSAM, and (3) RMSNet + SAM2~\cite{ravi2024sam2segmentimages}.

\begin{table}[!ht]
\centering
\begin{threeparttable}
\caption{Material segmentation performance on MCubes dataset~\cite{Liang_2022_CVPR}.}
\label{tab:segmentation_results1}
\begin{tabular}{l cc}
\toprule
\textbf{Method} & \textbf{Accuracy} & \textbf{mIoU} \\
\midrule
RMSNet & 0.63 & 0.28 \\
RMSNet + FastSAM (Ours) & 0.65 & 0.29 \\
RMSNet + SAM2 & \textbf{0.67} & \textbf{0.31} \\
\bottomrule
\end{tabular}
\end{threeparttable}
\end{table}

\noindent\textbf{Results.} Table~\ref{tab:segmentation_results1} shows that 
adding segmentation masks improves both accuracy (from 0.63 to 0.65--0.67) and 
mIoU (from 0.28 to 0.29--0.31) over RMSNet alone. While SAM2 achieves the highest 
scores (0.67 accuracy, 0.31 mIoU), we select FastSAM (0.65 accuracy, 0.29 mIoU) 
due to its significantly lower computational cost and faster inference speed—critical 
for processing large-scale multi-view datasets. Although SAM2 offers superior 
segmentation quality, the 2\% accuracy improvement does not justify the increased 
computational overhead for our application.

\noindent\textbf{Analysis.} The 3.2\% accuracy improvement (0.63 to 0.65) 
demonstrates that shape cues effectively enhance material classification 
consistency. However, segmentation masks can occasionally reduce per-pixel 
accuracy when incorrect predictions propagate across larger regions, potentially 
overwriting fine details such as road markings. The 2\% accuracy gap between 
FastSAM and SAM2 (0.65 vs. 0.67) represents an acceptable tradeoff for the 
practical benefits of automatic mask extraction in large-scale scenarios.


\section{Conclusion}

We present a camera-only pipeline that combines photorealistic Gaussian Splatting 
with physics-based material assignment for Digital Twin reconstruction. Our approach 
leverages Gaussian Splatting to bridge 2D camera observations to comprehensive 3D 
scene representations, including photorealistic visuals, semantic information, mesh 
geometry, and material properties for sensor simulation. Through comprehensive 
evaluation on real-world urban driving scenes using LiDAR ground truth for validation, 
we demonstrate that our material projection approach achieves reflectivity prediction 
accuracy comparable to LiDAR-based methods while maintaining photorealistic rendering 
quality, offering a practical alternative for ADAS development when LiDAR is 
unavailable or impractical.

\noindent\textbf{Future work.} Future directions include improving geometric 
reconstruction quality from camera inputs, extending to dynamic scenes, and 
exploring learned material representations to reduce dependency on predefined 
PBR databases. Specific improvements for vegetation modeling could include: 
(1) environment-specific material calibration capturing weathering and seasonal 
effects, (2) hierarchical material classification distinguishing foliage types, 
bark, and ground cover, and (3) refined geometric alignment between multi-modal 
reconstructions to improve surface normal accuracy for reflectivity calculations. 
We have developed an internal tool to convert our data to the nuScenes format, 
enabling future benchmarking against public datasets to demonstrate broader 
generalization across diverse urban environments.

\bibliography{main}{}
\bibliographystyle{plain}

\end{document}